\pdfoutput=1

\documentclass[11pt]{article}
\usepackage[]{ACL2023}
\usepackage[T1]{fontenc}
\usepackage[utf8]{inputenc}
\usepackage{times}
\usepackage{mathtools}
\usepackage{makecell}
\usepackage{booktabs}
\usepackage{amsfonts, amssymb}
\usepackage{graphicx}
\usepackage{latexsym}
\usepackage{tcolorbox}
\usepackage{multirow}
\usepackage{xcolor}
\usepackage{microtype}

\usepackage{inconsolata}
\usepackage{soul}

\title{When does word order matter and when doesn't it?}

\author{Xuanda Chen \\
  McGill University \\
  Mila Quebec AI Institute \\
  \texttt{xuanda.chen@mail.mcgill.ca} \\\And
  Timothy O'Donnell \\
  McGill University \\
  Mila Quebec AI Institute \\
  Canada CIFAR AI Chair \\
  \texttt{timothy.odonnell@mcgill.ca} \\\And
  Siva Reddy \\
  McGill University \\
  Mila Quebec AI Institute \\
  Canada CIFAR AI Chair \\
  \texttt{siva.reddy@mila.quebec} \\}

\begin{document}
\maketitle

\begin{abstract}

Language models (LMs) may appear insensitive to word order changes in natural language understanding (NLU) tasks. In this paper, we propose that linguistic redundancy can explain this phenomenon, whereby word order and other linguistic cues such as case markers provide overlapping and thus redundant information. Our hypothesis is that models exhibit insensitivity to word order when the order provides redundant information, and the degree of insensitivity varies across tasks. We quantify how informative word order is using mutual information (MI) between unscrambled and scrambled sentences. Our results show the effect that the less informative word order is, the more consistent the model's predictions are between unscrambled and scrambled sentences. We also find that the effect varies across tasks: for some tasks, like SST-2, LMs’ prediction is almost always consistent with the original one even if the Pointwise-MI (PMI) changes, while for others, like RTE, the consistency is near random when the PMI gets lower, i.e., word order is really important \footnote{The code can be found at https://github.com/xdchen2/order.git}.

\end{abstract}

\section{Introduction}

Language is a fundamental part of human communication, and its structure is highly compositional. This concept of compositionality is essential to language, because it allows us to process unseen sentences from the meaning of individual parts. For instance, in the sentence \textit{The cat chased the mouse}, the meaning of the sentence is derived from the meaning of \textit{cat}, \textit{chased}, and \textit{mouse} and the order. Languages require a specific word order. It is essential for humans to understand sentence meaning, and changing the order of the words can change the entire meaning of the sentence. As an example, consider the sentence above where altering the word order to \textit{The mouse chased the cat} would result in a sentence with an entirely opposite meaning.

\begin{figure}[tb]
    \centering
    \includegraphics[width=1\linewidth]{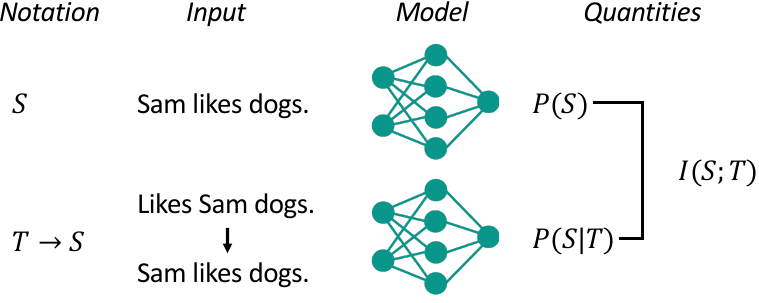}
    \caption{
        Variational approximation of the MI between scrambled and unscrambled sentences, using an LM and a reordering model. The estimation relies on bounding MI \textemdash see discussion in $\S$\ref{mi}.
    }
    \label{pc}
    \end{figure}

Despite the importance of word order in human language processing, recent studies have found that word order does not matter for many language models \citep[LMs;][]{sinha2021masked, sinha2021unnatural, pham2021out}. For example, natural language inference (NLI) is a task of determining whether the given “hypothesis” and “premise” logically follow (entailment) or unfollow (contradiction) or are undetermined (neutral) to each other. LMs can successfully identify the entailment relationship between the premise, \textit{A soccer game with multiple males playing} and the hypothesis, \textit{Some men are playing a sport}. \citet{sinha2021masked, sinha2021unnatural} found that LMs still predict correctly even when the premise is scrambled like \textit{males playing soccer game A with multiple} and the hypothesis like \textit{playing a are sport Some men}.

The findings of these studies raise questions about how LMs process language and whether word order is necessary for meaning computation. In the case of a sentence like \textit{The cat chased the mouse}, altering the word order can completely change the meaning. However, for a sentence such as \textit{The cat is chasing two balls}, the original meaning can still be understood even if the sentence is scrambled, as in \textit{Is chasing two balls the cat}. This is because the singular noun \textit{cat} is the only possible subject due to subject-verb agreement, and the animate subject requirement of the verb \textit{chase} further supports this inference. In these instances, both word order and grammatical agreement offer overlapping and thus redundant information, suggesting that word order may not always be indispensable in language processing.

In this paper, we propose the \textit{redundancy effect}, which suggests that LMs may not depend on word order in NLU tasks since it provides redundant information. To measure the redundancy, we consider scrambled and unscrambled sentences as two random variables and use mutual information (MI) to measure the average amount of information scrambled sentences contain about unscrambled sentences \citep{cover1999elements}. This measurement is agnostic to representation and processing, i.e., how LMs are trained and how scrambling is performed. We hypothesize that when the MI between scrambled and unscrambled sentences is high, word order is less crucial for LMs to solve NLU tasks. Unfortunately, MI is inaccessible, so we estimate it using a variational approximation method with a reordering model (RM) and an LM \citep{li2019specializing}. 

Our findings reveal that LMs face difficulty in handling certain scrambled sentences with low Pointwise-MI (PMI). However, they can successfully solve most NLU tasks with sentences that are easier to reorder. We also find that the size of the redundancy effect varies across tasks, indicating their different levels of sensitivity to word order. For some tasks such as SST-2, word order doesn't matter. Even if the PMI changes, the prediction is almost always consistent with the original one. For some tasks such as RTE, the prediction consistency depends on the PMI, that is, lower PMI sentences lead to inconsistent performance.

\section{MI Estimation}
\label{mi}
Mutual information (MI) is a powerful tool for measuring the statistical dependence between random variables, and here, the dependence between scrambled and unscrambled sentences. Let $S$ be a random variable that ranges over possible sentences as strings of words, and $P(S)$ be the distribution over all sentences. Let $T$ be a random variable that ranges over all possible scramblings of $S$, and let $\sigma(\cdot)$ be the scrambling function, such that $p(t) = \underset{s \in \sigma(t)}{\Sigma} p(s) \frac{1}{|s|!}$. The MI between unscrambled and scrambled sentences can be denoted as $I(S;T) \coloneqq \underset{s,t \sim p(s,t)}{\mathbb{E}} \left[ \log \frac{p(s|t)}{p(s)} \right]$. This is the expected value of the logarithmic ratio between the probability of sentences given scrambling and the probability of sentences for all possible sentence and scrambling pairs. In practice, estimating MI can be challenging because we only have access to samples, not the underlying distributions. To establish an initial lower bound on MI, we replace the intractable conditional distribution $p(s|t)$ with a tractable optimization problem over a variational distribution $q_{\phi}(s|t)$. This allows us to obtain a lower bound on MI due to the non-negativity of the KL divergence,

\begin{equation} \label{eqx}
\begin{split}
I(S;T) & = \underset{s \sim p(s)}{\mathbb{E}} \underset{t|s \sim p(t|s)}{\mathbb{E}} \left[ \log \frac{q_{\phi}(s|t)}{p(s)} \right] \\
 & + \underset{t \sim p(t)}{\mathbb{E}} \left[ KL(p(s|t)\vert\vert q_{\phi}(s|t)) \right] \\
\end{split}
\end{equation}

Given the fixed value of MI and the non-negativity of KL, maximizing our estimate is equivalent to minimizing KL by finding a $q_{\phi}$ that approximates the true distribution $p$ well. We define $q_{\phi}$ as the reordering model (RM) tasked with restoring the original word order from scrambling. The RM is trained on the sentence-scrambling pairs, $(s,t)$, from a corpus where each sentence corresponds to several possible scramblings. The optimal $q_{\phi}$ is the one that minimizes KL and maximizes the expected value of $\log q_{\phi}(s|t)$ over all $(s,t)$ pairs. This means that finding the optimal probability assignment $q_{\phi}$ on average is equivalent to finding the $q_{\phi}$ that best matches $p(s|t)$. We can use the lower bound as an approximation for the PMI of each sentence, which is calculated as $\operatorname{pmi}(s;t)=\log_{2} \frac{q_{\phi}(s|t)}{p(s)}$. Here, $q_{\phi}(s|t)$ is estimated by the RM, and $p(s)$ is provided by a pre-trained LM, T5 \citep{raffel2020exploring}.


\section{Training the Reordering Model}

We present the training procedure used for the RM. We repurposed the T5 model as the RM to estimate $q_{\phi}(s|t)$. We created a dataset by sampling 100,000 sentences from English Wikipedia, with each sentence scrambled at the unigram word level using six different random seeds, resulting in 600,000 $(s,t)$ pairs. We split this dataset into 90\% for training and 10\% for validation. The RM was trained for 10 epochs with a learning rate of 1e-4, taking approximately 5.6 hours to train on a single GPU.

\section{Validating the Reordering Model}

We further investigate the ability of the RM to infer original word order by exploiting other linguistic cues such as grammatical agreement and animacy. To assess this ability, we conducted an evaluation of the RM on a novel dataset that was not used during training. The dataset comprises two sentence types, referred to as type A and type B, which differ in their argument structures. In type A sentences, subjecthood/objecthood can be inferred through the grammatical agreement or animacy requirements of the verbs, as exemplified by the sentence "Sam throws the rock." Conversely, in type B sentences, such as "Sam beats John," word order is crucial in determining sentence meaning, as neither grammatical agreement nor animacy alone can identify the subject and object.

We hypothesize that the RM should perform well in reconstructing the original word order for type A sentences due to the presence of linguistic cues. In contrast, for type B sentences, where word order is essential, we expect the RM's accuracy to be lower. This is because, for instance, "Sam beats John" and "John beats Sam" are equally probable. We created two datasets of 1000 sentences each using a context-free grammar (CFG). We then scrambled the sentences across words using six random seeds. We computed the average accuracy of reordering based on the six random scramblings. Our results show that the RM achieved an accuracy of 0.948 for type A sentences and 0.506 for type B sentences. This confirms that the RM can leverage other linguistic cues to compensate for missing word order information and successfully reconstruct it.

\section{Experiment, Data and Results}

Recent studies have highlighted the insensitivity of LMs to word order in NLU tasks. However, it remains unclear whether word order is dispensable in all contexts across different tasks and sentences. As proposed earlier, the \textit{redundancy effect} posits that LMs may not rely on word order when it provides redundant information. We employed a Bayesian mixed-effect logistic model to explore the relationship between word order redundancy and LMs' performance in NLU tasks.

The predictor variable, word order redundancy, was quantified as the average PMI between an unscrambled sentence and its all possible scramblings. We performed random sampling from all possible scramblings to approximate the average PMI. Specifically, we scrambled each sentence across words with six random seeds. The response variable, \textsc{consistency}, showed whether the LM output the same label on an NLU task for both scrambled and unscrambled sentences. A score of 1 indicated consistent predictions for scrambled and unscrambled inputs, while a score of 0 indicated inconsistency. We also included \textsc{sentence length} as a confounding predictor to account for its impact on PMI estimation. Our regression model considered both random intercepts and slopes unique to each task. Random intercepts captured task difficulty, with higher values indicating simpler tasks that could be solved using a bag-of-words approach. In contrast, random slopes reflected the redundancy effect specific to each task, with larger values indicating greater sensitivity to word order scrambling.

Consistent with \citet{sinha2021masked}, we tested RoBERTa, a Masked LM proposed by \citet{liu2019roberta}, and assessed its performance across a diverse range of binary classification tasks from popular benchmarks:

\begin{description}
   \item[SST-2] \citep{socher2013recursive} a sentiment analysis task to analyze one-sentence movie reviews on the emotional tone as positive or negative.
   \item[MRPC] \citep{dolan2005automatically} a sentence similarity task to determine whether pairs of sentences are semantically equivalent or not.
   \item[QQP] \citep[GLUE;][]{wang2019glue}  a sentence similarity task to determine whether a pair of questions asked on Quora is semantically equivalent.
   \item[RTE] \citep[GLUE;][]{wang2019glue} an inference task to determine whether a given statement (the hypothesis) can be inferred from a given text (the premise), either as `entailed' or `not entailed'.
   \item[COPA] \citep{roemmele2011choice} an inference task, where a model is presented with a premise sentence and a query with two alternatives, and the model must choose the alternative that is most logically related to the premise. 
   \item[BoolQ] \citep{clark2019boolq} a QA task to answer multiple binary questions based on a given passage of text. 
   \item[WinoGrande] \citep{sakaguchi2021winogrande} a commonsense reasoning task to answer multiple-choice questions that require understanding of how events in the world typically unfold based on common sense knowledge.
\end{description}

\begin{figure}[h]
    \centering
    \includegraphics[width=0.95\linewidth]{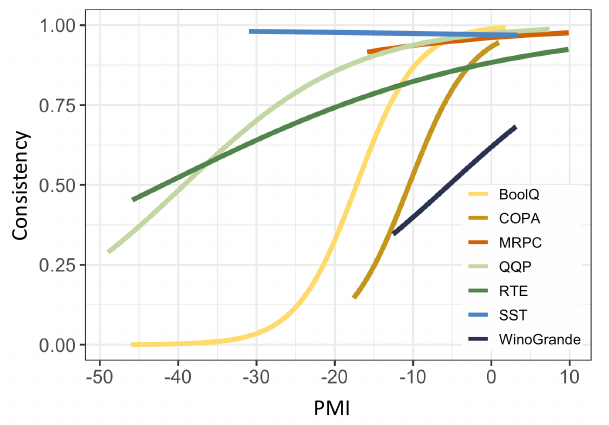}
    \caption{
        Curves represent a simulation of the linear model for the \textit{redundancy effect}. The x-axis reflects the actual data range, and the displayed ranges differ for each task. Line slopes indicate the level of redundancy effect, where steeper lines reflect a more pronounced effect. The intercept (the baseline level of the PMI influence on Consistency, i.e., how different tasks might have different starting points) signifies task difficulty: tasks with lower intercepts are more challenging for LMs to solve.
    }
    \label{pc2}
    \end{figure}

\subsection{Regression Results} 
To evaluate the redundancy effect of word order on LMs in NLU tasks, we limited our analysis to test samples where RoBERTa produced accurate predictions. This is because LMs can make inaccurate predictions for various reasons, which can confound the analysis. Our model revealed a significant redundancy effect at the group level. Specifically, the PMI strongly predicted the consistency of RoBERTa's predictions ($\hat{\beta}$=1.87, SE=0.91, CI=[-0.09,3.67], ROPE$>$0.999). We further investigated the variability in the redundancy effect across tasks by plotting the fitted curves for each task in Figure~\ref{pc2}. The x-axis represents PMI, which quantifies word order redundancy, and the y-axis represents consistency as a measure of LMs' performance. Our results showed significant differences among the tasks in both intercepts and slopes. For instance, the tasks SST-2 and MRPC exhibited almost flat curves, implying that word order was not critical to these tasks. Conversely, the tasks BoolQ and COPA showed steeper curves, indicating that they were highly sensitive to word order scrambling, with the LM's performance declining rapidly when the word order was scrambled. Furthermore, our fitting revealed that some sentences in BoolQ, QQP, and RTE with low PMI were more challenging to reconstruct when word order was not available. In conclusion, our findings suggest that LMs can handle scrambled sentences when the original word order is recoverable, but the degree of sensitivity varies depending on the task. 

\begin{table*}[ht]
    \small
    \centering
    \begin{tabular}{p{2.9in}p{2.9in}}
    \toprule
    \textbf{Original Sentences}  & \textbf{Generated Sentences} \\ \midrule
    what would be the effect on India if Donald Trump really becomes the president of US? & if the president of the India becomes really Donald Trump what would be the effect on US?            \\ [0.8ex]
    how much time I need to wait for the result of South Korean student visa?       & South Korean student wait for the result of visa how much time I need to?        \\ [0.8ex]
    consumers who said jobs are difficult to find jumped to 32        & 32 consumers who are said to be difficult to find jobs jumped         \\ [0.8ex]
    Germany is the powerhouse of the EU         & the powerhouse of the EU is Germany         \\ [0.8ex]
    why were the Viking raids so successful?    & the Viking raids were successful why so? \\ [0.8ex]
    \bottomrule
    \end{tabular}
    \caption{Examples of probe model generation.}
    \label{tabx}
    \end{table*}

\subsection{Case Studies on Negative-PMI Sentences} 
We observed that a considerable number of sentences have low negative PMI values, indicating that the probability of a sentence given its scrambling is much lower than the probability of that sentence. It suggests that the LM believes there are better orders than the original word order. In Table~\ref{tabx}, we provide several examples of such reconstructed sentences that were manually checked. While some reconstructed sentences are grammatically correct, they may result in hilarious or nonsensical meanings. For instance, one original sentence discussed the potential impact of Donald Trump's presidency on India, but the generated sentence asked about the impact on the United States if Donald Trump became the President of India. In another instance, a sentence's intended meaning was that the number of consumers had decreased to 32, but the recovered meaning suggested that consumers were committing suicide because they could not find jobs. Moreover, these examples demonstrate the LM's strong inductive bias towards declarative sentences. For instance, in the last example, the LM transformed a statement into a question by using the phrase `why so,' despite the original sentence being a standard question.

\section{Discussion}

This study aims to address when word order is and is not dispensable for LMs to achieve NLU tasks. We introduce the concept of the redundancy effect and demonstrate that word order may not be critical when it provides redundant information that can be substituted by other linguistic cues. We observe that the redundancy effect varies across sentences, and some sentences are challenging to comprehend without the correct word order. This effect also differs among NLU tasks, with some tasks being more sensitive to word order than others, such as COPA. In addition, we conduct case studies on sentences with low negative PMI and find that they could be reordered in different ways that are both grammatically correct and semantically sensible.

Recent studies have shown that humans have sophisticated interpretation mechanisms that enable them to ignore certain form-based errors in scrambled word order \citep{traxler2014trends, mirault2018you}. When the word order is scrambled in specific locations, humans can still parse the sentence and derive meaning \citep{mollica2020composition}. We believe that our concept of the redundancy effect could also account for this observation: the part of word order that has been scrambled does not provide additional information beyond what other linguistic cues already offer within the sentence.

\section{Related Work}

There is a growing body of research exploring the insensitivity of LMs to word order scrambling. To start with, \citet{sinha2021unnatural} found that LMs can achieve NLI with high accuracy even when sentences in the dataset are scrambled. Similarly, \citet{ettinger2020bert} analyzed LMs' performance on several NLI tasks and concluded that LMs may overattend to individual words and ignore structural information, which may explain their insensitivity to word order. To further test the effect of training schemes, \citet{sinha2021masked, gupta2021bert} re-trained LMs with scrambled sentences and found that these LMs can still solve NLI tasks without a drop in performance. Moreover, \citet{pham2021out} investigated how this phenomenon varies across different NLU tasks. They tested several other NLU tasks from the GLUE \citep{wang2019glue} and SuperGLUE \citep{wang2019superglue} benchmarks and found that some tasks are more sensitive to word order scrambling than others, such as CoLA and RTE. While these studies focus on word-level scrambling, \citet{clouatre2022local} analyzed finer-grained scrambling at the subword and character levels. They found that LMs, regardless of their inductive biases, pretraining scheme, or the choice of tokenization, can achieve NLU tasks when local structures are kept unscrambled and LMs make limited use of global structure. In contrast to these studies, our focus is on how sentences differ in their impact on LMs' performance after scrambling, and our proposed method is agnostic to representation and processing, i.e., the way the LMs are trained and the scrambling is performed.

\section{Conclusion}

In this paper, we provide computational evidence to support the claim that linguistic redundancy can explain LMs' insensitivity to word order. Despite unexplained variance, our regression model shows a strong redundancy effect across all tasks, indicating that LMs accept scrambled texts when word order is less informative and recoverable. We also find that the effect varies across tasks: for some tasks, like SST-2, LMs’ prediction is almost always consistent with the original one even if the PMI changes, while for others, like RTE, the consistency is near random when the PMI gets lower, which shows that word order is really important in this case. Future research could include more tasks and adopt more powerful models for MI estimation.

\bibliography{lit}
\bibliographystyle{acl_natbib}

\appendix

\section{Regression Model}

\subsection{Generalized Linear Model}

Throughout our experiments, we utilize Bayesian logistic regression models. To begin, let us recall the definition of the classical linear regression model. Given a training set $\mathcal{D}$ consisting of inputs and targets ${(\mathbf{x}^{(n)}, \mathbf{y}^{(n)})}_{n=1}^{N}$, the regression model is defined as follows:

\begin{equation}
\label{1}
\mathbf{y}=\mathbf{w}^\top \mathbf{x} + \epsilon, \epsilon \sim \mathcal{N}(0,\sigma^2_{err}) \
\end{equation}

Here, $\mathbf{y}$ represents the target variable, $\mathbf{x}$ represents the input variable, and $\mathbf{w}$ represents the learned weights that reflect the degree to which $\mathbf{y}$ is conditioned on $\mathbf{x}$. The error term $\epsilon$ represents the unexplained variance in $\mathbf{y}$. It is important to note that logistic regression differs from standard linear regression in the way that the probability of a particular outcome is linked to the linear predictor function. Specifically, the logit function, which is the natural logarithm of the odds, is used to convert the probability (which is bounded between 0 and 1) to a variable ranging over $(-\infty, +\infty)$. This transformation is used to match the range of the linear prediction function.

Maximum Likelihood Estimation (MLE) is often used to estimate the weight variable $\mathbf{w}$. The likelihood of the model is derived, and then maximized with respect to $\mathbf{w}$ using an optimization algorithm such as Gradient Descent. However, it is important to note that MLE assumes that the data is independently and identically distributed (i.i.d.). This assumption is not always satisfied, and the estimation can therefore be unreliable. To address this issue, we use Maximum A Posteriori (MAP) estimation, which is defined under the Bayesian framework and works on a posterior distribution (as opposed to the likelihood alone). The inclusion of the prior $P(\mathbf{w})$ in MAP leads to more robust estimation of parameters. In our case, we do not have balanced data points across different tasks, and some tasks have only a few samples. Bayesian regression helps us to make more robust comparisons between the estimated $\mathbf{w}$ across tasks by sampling from the actual shape of the posterior distribution and obtaining confidence intervals.

\subsection{Generalized Mixed Effects}

In our study, co-task samples are expected to exhibit dependencies in how the error terms are sampled. To model these dependencies through random effects, we adopt mixed-effects models. Specifically, for input vector $\mathbf{x}$ from a particular task $k$, our model is defined as follows:

\begin{equation}
\label{3}
\mathbf{y}=\mathbf{w}^\top \mathbf{x} + \mathbf{r}_{k} + \epsilon \
\end{equation}

Here, $\mathbf{y}$ denotes the target variable, $\mathbf{w}$ is the learned weight vector that captures the degree to which $\mathbf{y}$ is dependent on $\mathbf{x}$, and $\epsilon$ represents the unexplained variance in $\mathbf{y}$. In addition, the random intercept $\mathbf{r}$ reflects the individual differences in the mean across all conditions and is assumed to be shared across all samples in task $k$.

To estimate individual differences in the effect of a predictor, i.e., different $\mathbf{w}$ for each task, we can add random slopes to the model. The updated formulation is as follows:

\begin{equation}
\label{4}
\mathbf{y}=\mathbf{w}^\top \mathbf{x} + \mathbf{s}{k}^\top \mathbf{x} + \mathbf{r}{k} + \epsilon \
\end{equation}

Here, each $\mathbf{s}{k} \sim \mathcal{N}(0,\Sigma\mathbf{s})$ represents a task-specific random slope, and $\Sigma_\mathbf{s}$ is a learned covariance matrix that shows the variance of $\mathbf{s}$ across tasks. We interpret random intercepts as the baselined task difficulty and random slopes as the sensitivity to word order.

\section{Response Variable and Predictors}

\subsection{consistency as response} 

We aim to investigate whether the LMs can accurately predict the meaning of scrambled texts, comparable to the predictions it makes on original texts. To this end, we define an \textit{consistency} that measures the consistency between the LM's predictions on scrambled texts and the corresponding predictions on the original texts. If the predictions on scrambled texts and normal texts are consistent, we assign a value of 1 to the consistency; otherwise, we assign a value of 0.

To carry out this investigation, we utilize RoBERTa \citep{liu2019roberta} as our prediction model and focus on test samples where it correctly predicts all the original texts. We then filter a new dataset, denoted as $\mathcal{D} = {(x,y) \rvert f(x) = y}$, where $f(x)$ represents the LM's prediction on text input $x$. We use this dataset to train a regression model that models the consistency as a function of various predictors.

\subsection{PMI and Sentence length as predictors}

We aim to investigate the effect of point-wise mutual information (PMI) on the consistency of scrambled texts, with the goal of shedding light on the redundancy effect. As such, PMI is included as a predictor in our regression model. Additionally, it is important to consider sentence length as a potential confounder, as it may also influence the consistency. However, including both predictors in the model may result in collinearity and unreliable coefficient estimation.

\begin{figure}[ht!]
    \centering
    \includegraphics[width=0.95\linewidth]{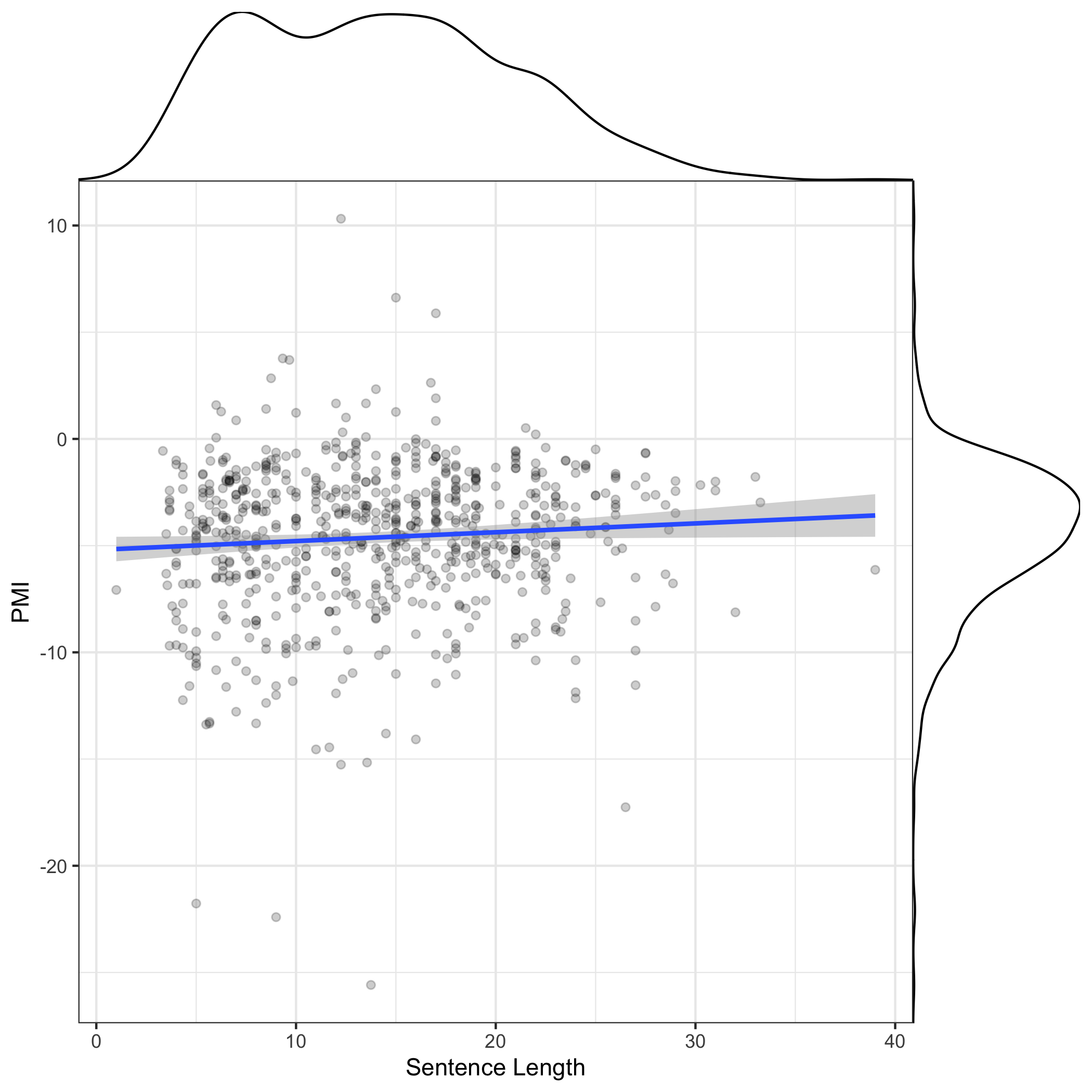}
    \caption{
        Correlation between sentence length and PMI.
    }
    \label{len}
    \end{figure}

To assess the relationship between sentence length and PMI, we conducted a correlation analysis, as shown in Figure~\ref{len}. Our findings suggest that while there is some correlation between the two variables, it is not substantial.

To address the potential collinearity, we fitted two regression models, one with sentence length as a predictor and one without. Based on Bayes Factors, the model including sentence length was found to better fit the data, and thus we have chosen to report statistics based on this model.

\subsection{Redundancy Effect at Group and Individual Level}

\begin{figure}[!ht]
    \centering
    \includegraphics[width=0.95\linewidth]{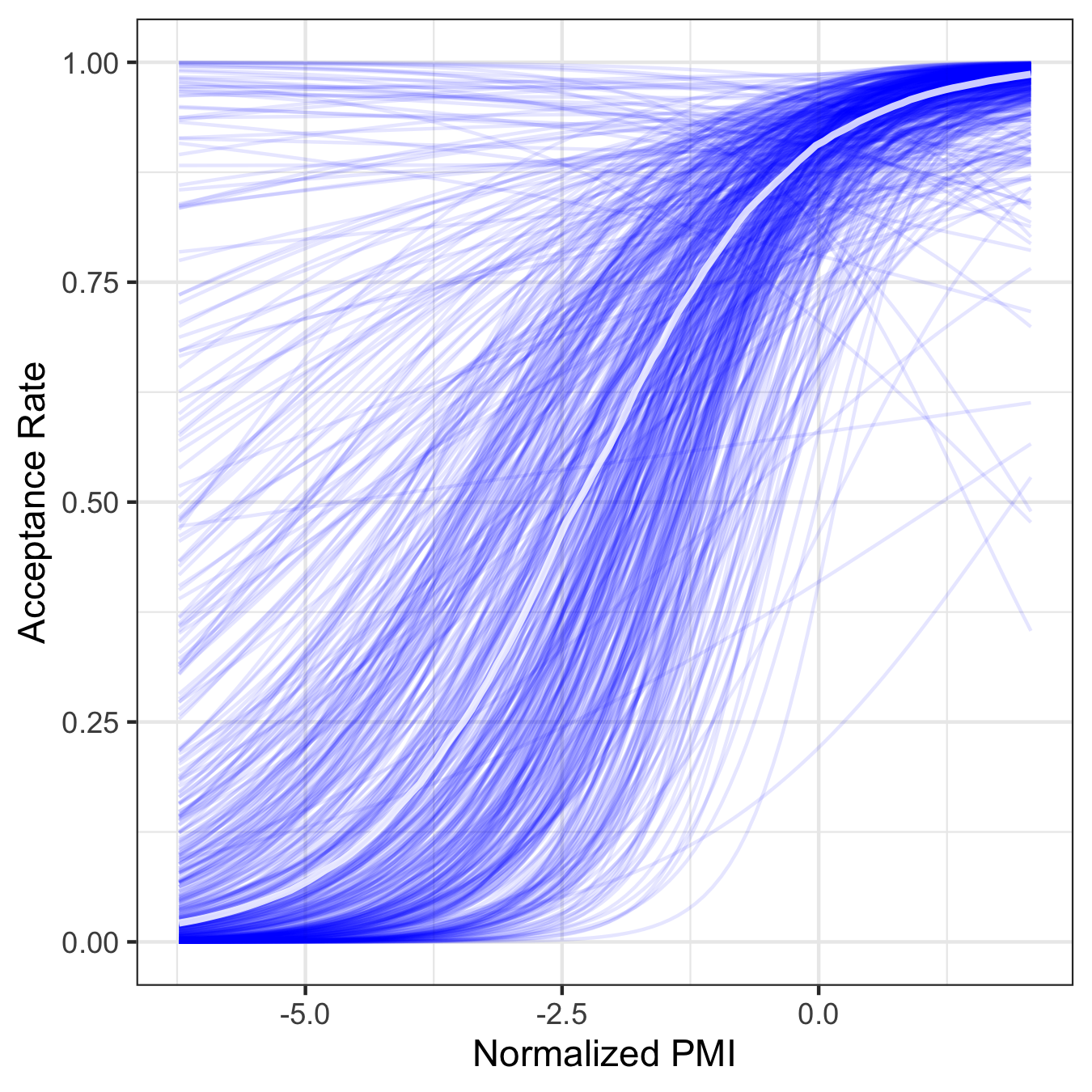}
    \caption{
        Conditional effect of PMI on consistency. The effect is estimated on random posterior draws from the model.
    }
    \label{pmi}
    \end{figure}

\begin{figure}[!ht]
    \centering
    \includegraphics[width=0.95\linewidth]{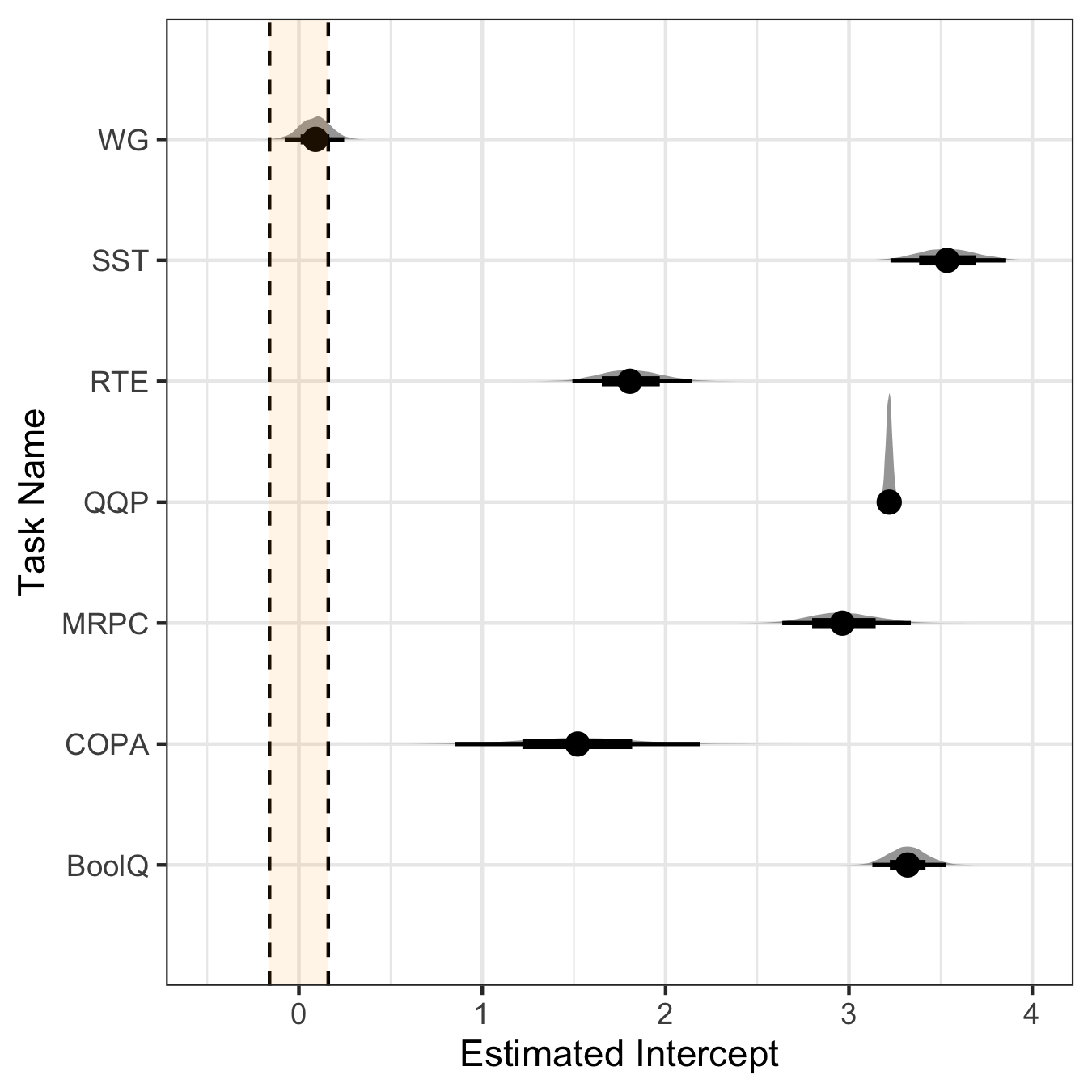}
    \caption{
        Estimated random intercepts across all tasks. The shaded area is the ROPE showing whether the effect size is big enough to be effective.
    }
    \label{re2}
    \end{figure}

In Figure~\ref{re2}, we present the individual differences in the mean across all conditions, represented by the random intercepts. A higher mean indicates a potential ceiling effect, whereby the baselined consistency remains high regardless of the sentence PMI. We sought to determine the statistical significance of these estimates, but Bayesian inference does not rely on statistical significance. Instead, we interpret the probability distribution, and values falling outside a predefined range are considered to have no practical effect or a negligible magnitude. This range is referred to as the region of practical equivalence (ROPE), and we set it at a range of -0.18 to 0.18 of a standardized parameter, which corresponds to a negligible effect size according to \citet{mcelreath2020statistical}. Our results indicate that all estimations, except for WinoGrande, are practically effective, indicating that all tasks, except WinoGrande, accept scrambled inputs significantly above chance.

\begin{figure}[!ht]
    \centering
    \includegraphics[width=0.95\linewidth]{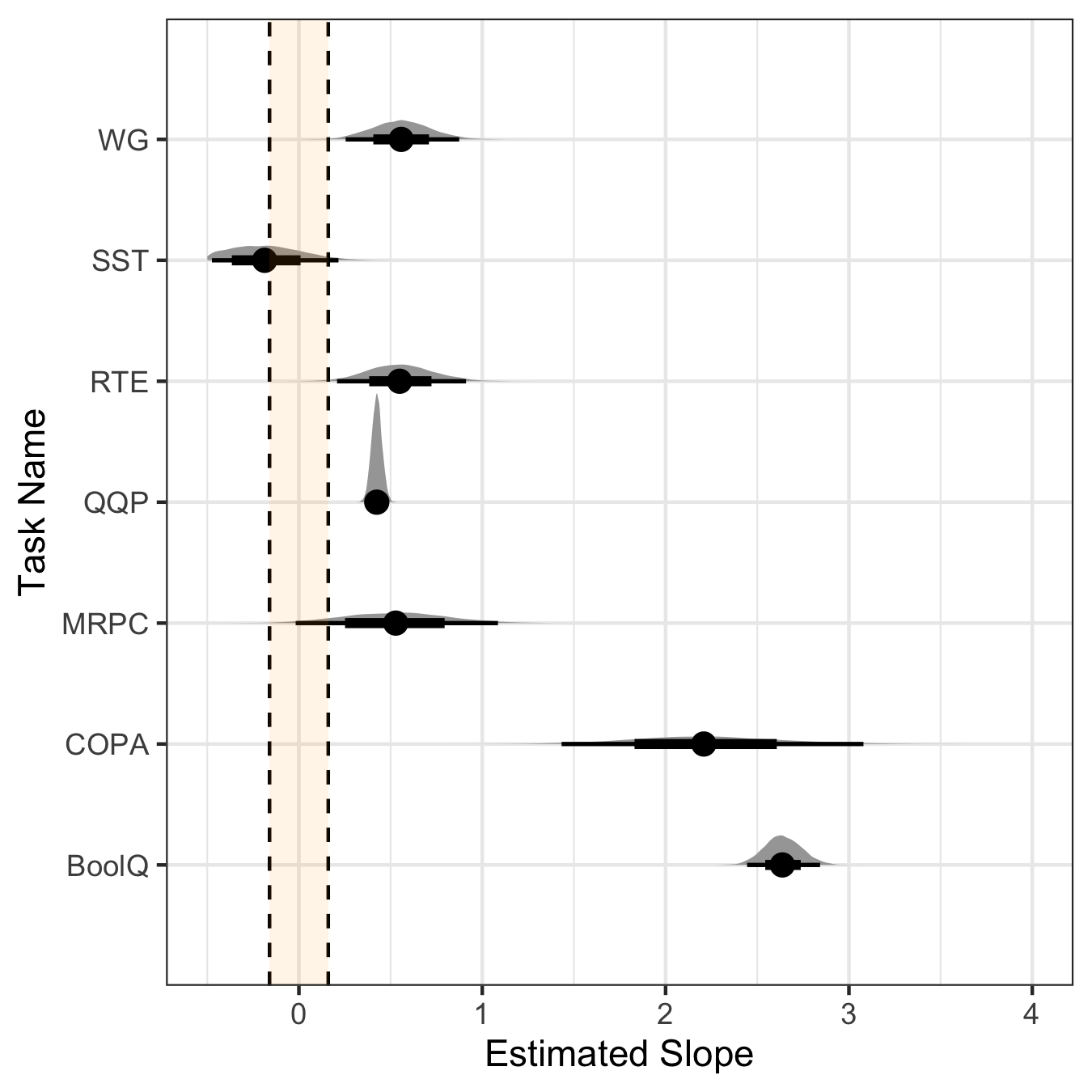}
    \caption{
        Estimated random slopes of PMI.
    }
    \label{re1}
    \end{figure}

Figure~\ref{re1} illustrates variations in the effect of PMI across tasks, which reflects the variance in sensitivity to word order. While SST-2 does not exhibit a notable effect, all other tasks show varying degrees of sensitivity, with BoolQ and COPA demonstrating the strongest.

\section{Reordering Model Evaluation}
\label{sec:appendix}

\subsection{Probe accuracy on the CFG datasets}

We endeavored to investigate the ability of a probe model to restore correct word order from perturbation, specifically focusing on two distinct types of argument structures - common noun and proper noun. To this end, we constructed two datasets using context-free grammar (CFG), each containing 1000 sentences with varying numbers of tokens, and subsequently scrambled them randomly six times, to evaluate the performance of the reordering model on a range of perturbations.

\begin{figure}[!ht]
    \centering
    \includegraphics[width=0.95\linewidth]{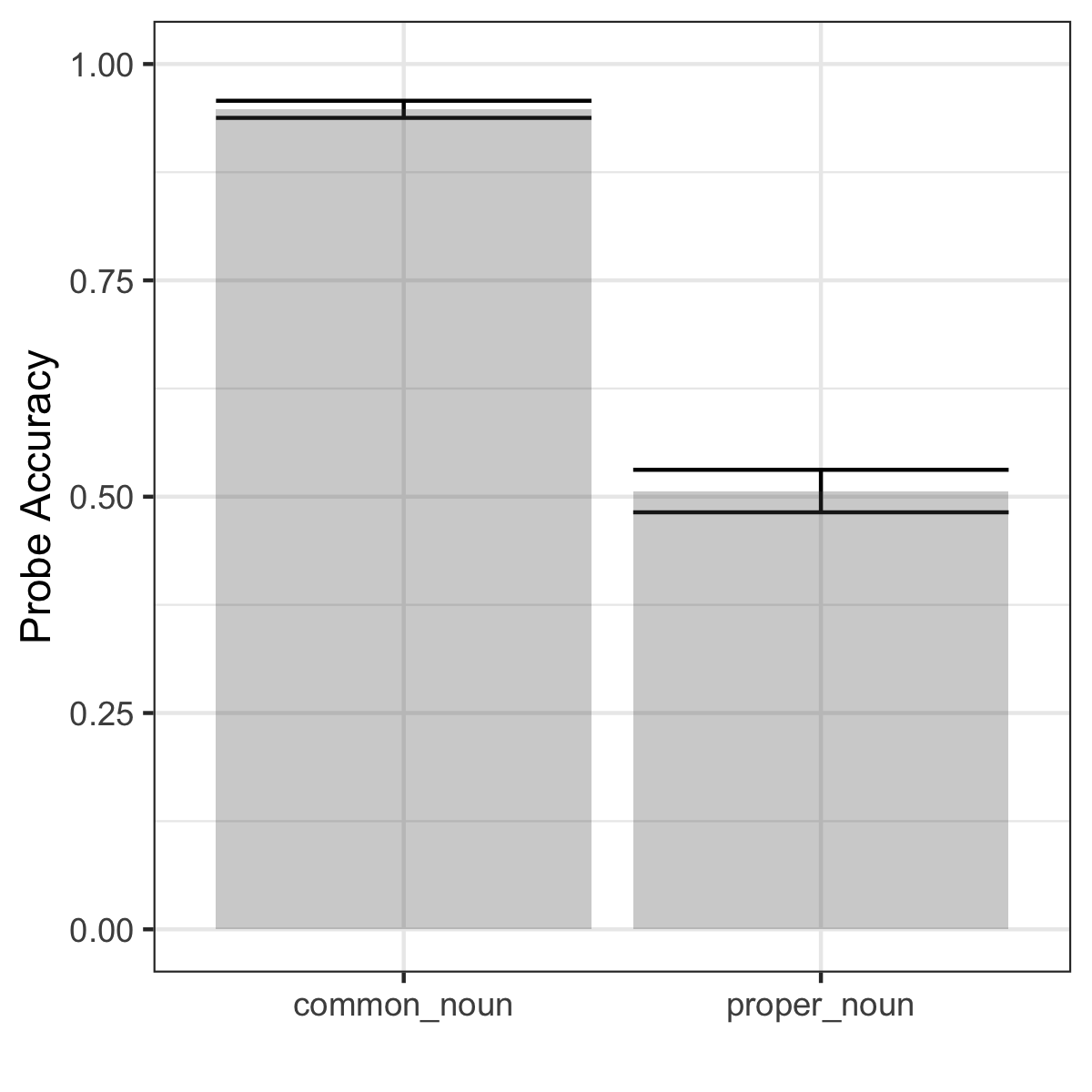}
    \caption{
        Probe accuracy on the CFG dataset.
    }
    \label{probe-acc}
    \end{figure}

Our experimental results, presented in Figure~\ref{probe-acc}, reveal that the reordering model exhibits high accuracy (0.948) in restoring common noun structures, while demonstrating chance performance (0.506) on the proper noun structures. Although the variance within proper noun accuracy is slightly higher than that within common noun accuracy, overall variance is low.

\subsection{Goodness of Probe Model}

We aim to investigate the efficacy of the reordering model in restoring the original word order from perturbed sentences. To achieve this goal, we propose a metric based on the Levenshtein Distance (LD), which is a standard measure of the minimum number of single-character edits required to transform one sequence into another \citep{levenshtein1966binary, yujian2007normalized}. Specifically, we define the Probe Accuracy (PA) as $1-norm(LD(o,r))$, where $o$ represents unscrambled sentences while $r$ reconstructed sentences. To ensure methodological rigor, we include a control baseline that measures the perturbation degree. This Control Accuracy (CA) is computed as the same quantity between unscrambled and scrambled sentences.

\begin{figure}[!ht]
    \centering
    \includegraphics[width=0.95\linewidth]{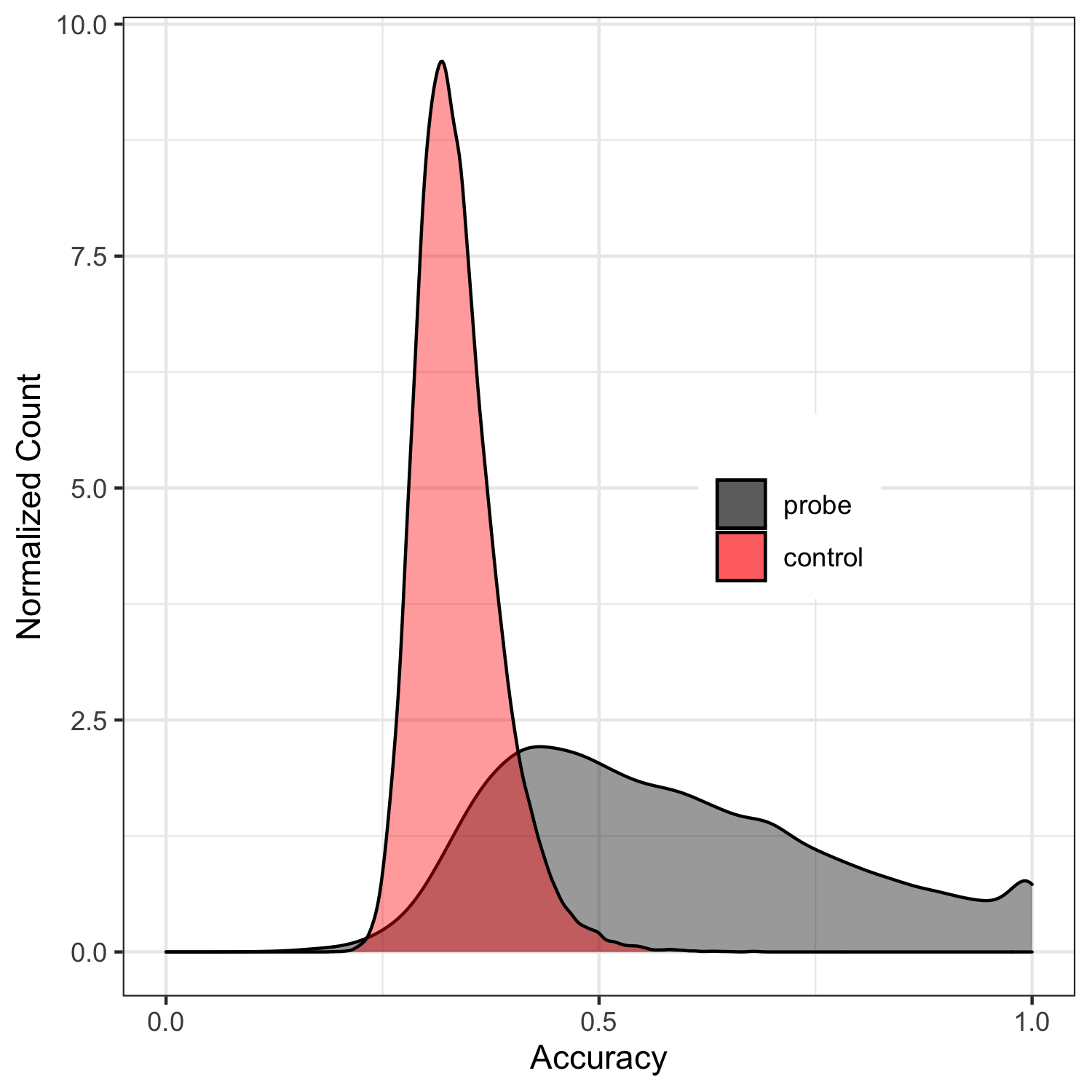}
    \caption{
        Distribution of PA and CA scores across tasks.
    }
    \label{da}
    \end{figure}

\begin{figure}[!ht]
    \centering
    \includegraphics[width=0.95\linewidth]{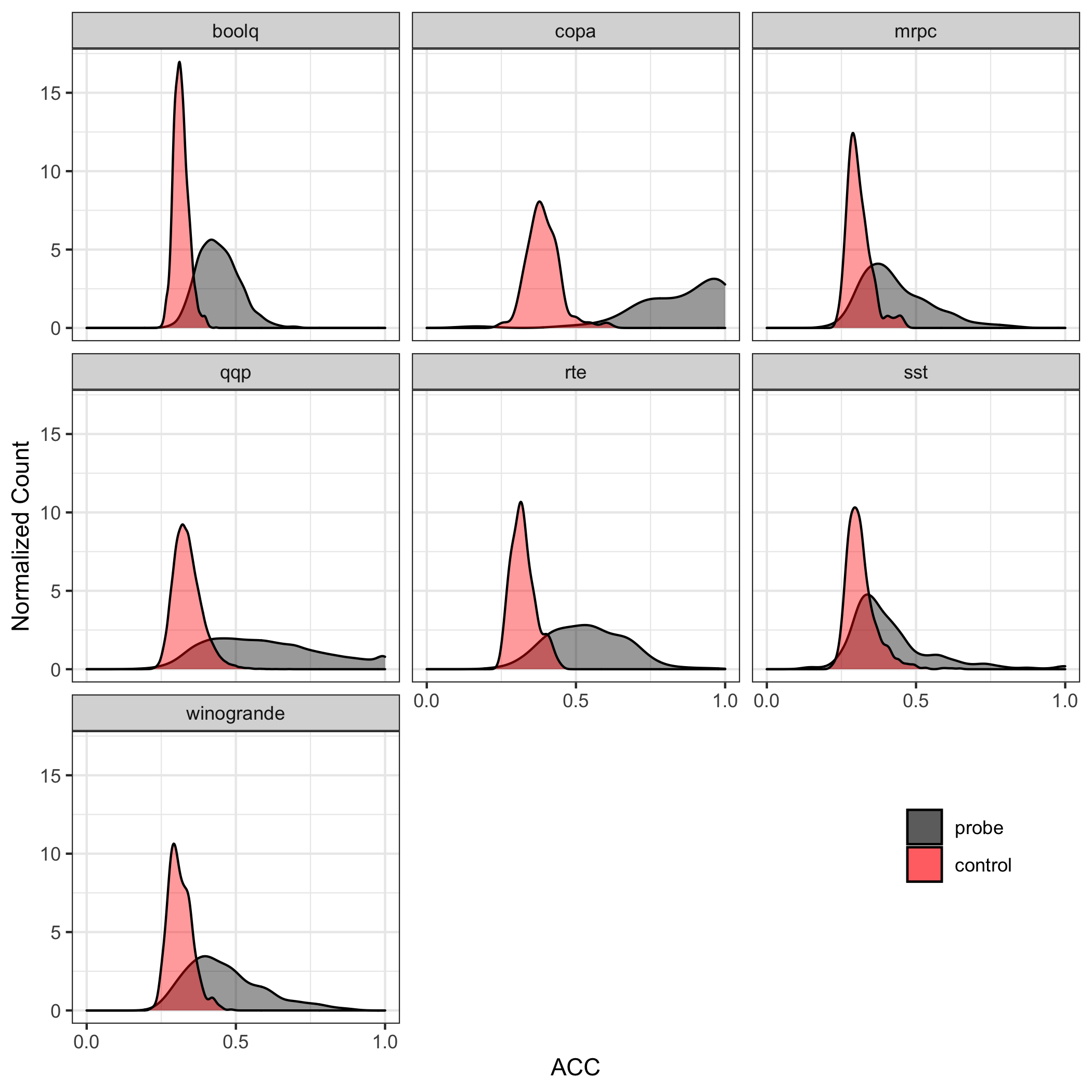}
    \caption{
        Distribution of sentence-level PA and CA scores across tasks.
    }
    \label{da-facet}
    \end{figure}

We present our findings on the PA and CA scores for all sentences in the tested tasks. As depicted in Figure~\ref{da}, the CA scores exhibit a highly dense distribution around 0.3, indicating that scrambling may leave some local structures unscathed. However, the PA scores flatten the distribution of the CA scores, demonstrating that the reordering model can restore more local word order, although the reconstruction may not be perfect. An overlapping area between the PA and CA scores suggests that some sentences are difficult to recover, and the reordering model may struggle with such sentences. A more detailed presentation of the results can be found in Figure~\ref{da-facet}.

\section{More information on datasets}

We have curated a selection of natural language understanding (NLU) tasks from popular benchmarks for our study. Table~\ref{tab1} provides a comprehensive overview of these datasets (including the number of sentence pairs from train and validation datasets, the average sentence length, and the number of sentences), highlighting their differences in terms of size and complexity. While some are considerably large, such as QQP, others are relatively small, such as COPA.

\begin{table}[ht]
    \small
    \centering
    \begin{tabular}{ccccc}
    \toprule
    \multirow{2}{*}{\textbf{Tasks}}  & \multicolumn{2}{c}{\textbf{Samples}} & \multicolumn{2}{c}{\textbf{Sent Stat}} \\ \cmidrule(l){2-3} \cmidrule(l){4-5}
                & Train  & Val   & \#Len  & \#Num    \\ \midrule
    SST-2       & 67349  & 872   & 16.912 & 915      \\ [0.6ex]
    MRPC        & 3668   & 408   & 17.512 & 1162     \\ [0.6ex]
    QQP         & 363846 & 40430 & 10.480 & 87635    \\ [0.6ex]
    RTE         & 2490   & 277   & 16.772 & 943      \\ [0.6ex]
    COPA        & 400    & 100   & 5.991  & 300      \\ [0.6ex]
    BoolQ       & 9427   & 3270  & 19.677 & 18614    \\ [0.6ex]
    WG          & 40398  & 1267  & 17.062 & 1515     \\ [0.6ex]
    \bottomrule
    \end{tabular}
    \caption{Statistics of selected NLU tasks (WG short for WinoGrande). Sentence statistics are calculated on the validation sets.}
    \label{tab1}
    \end{table}

\begin{figure}[!ht]
    \centering
    \includegraphics[width=0.95\linewidth]{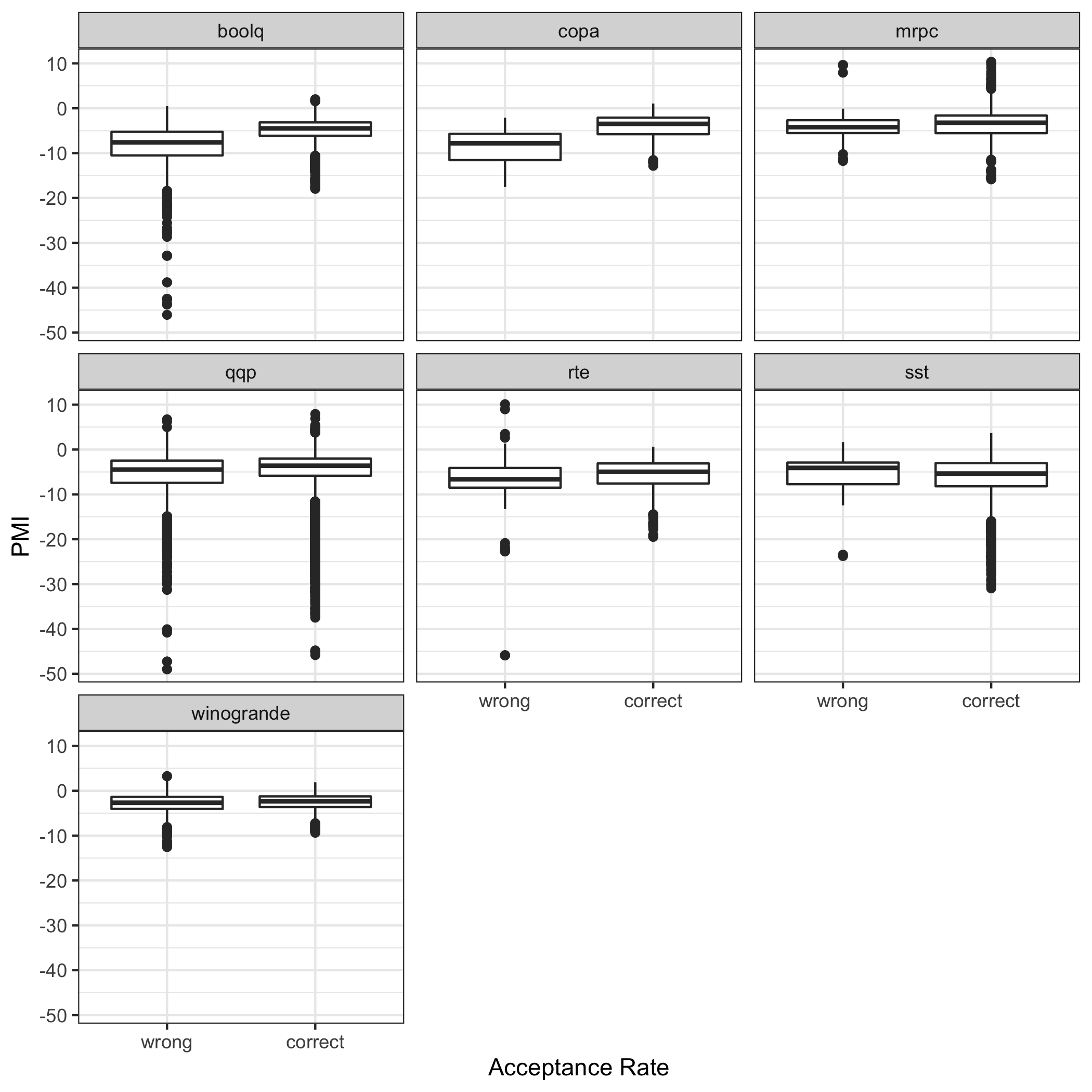}
    \caption{
        Boxplots of PMI distribution across tasks.
    }
    \label{box}
    \end{figure}

To gain a more in-depth understanding of the task design, we have included specific examples from each of these tasks, illustrating both the input and output components in Table~\ref{box}. 

We note that the difficulty of reconstructing word order from scrambling varies across tasks. Figure~\ref{box} showcases the distribution of by-task PMI scores, where higher PMI scores generally imply that the original word order is easier to restore.

\end{document}